\pdfoutput=1

\documentclass[11pt]{article}

\usepackage{acl}

\usepackage{times}
\usepackage{latexsym}
\usepackage{amsfonts}
\usepackage{amsmath}
\usepackage{amssymb}
\usepackage{hyperref}
\usepackage{multirow}
\usepackage{xcolor}

\usepackage[noabbrev,capitalize]{cleveref} %

\crefname{footnote}{footnote}{footnotes}   %
\crefname{line}{line}{lines}               %
\crefname{equation}{equation}{equations}   %
\crefname{section}{\S}{\S\S}
\Crefname{section}{\S}{\S\S}    %
\crefformat{section}{#2\S#1#3}  %
\Crefformat{section}{#2\S#1#3}
\crefrangeformat{section}{\S\S#3#1#4--#5#2#6}
\Crefrangeformat{section}{\S\S#3#1#4--#5#2#6}
\crefmultiformat{section}{\S#2#1#3}{ and~\S#2#1#3}{, \S#2#1#3}{ and~\S#2#1#3}
\Crefmultiformat{section}{\S#2#1#3}{ and~\S#2#1#3}{, \S#2#1#3}{ and~\S#2#1#3}
\crefrangemultiformat{section}{\S\S#3#1#4--#5#2#6}{ and~\S\S#3#1#4--#5#2#6}{, \S\S#3#1#4--#5#2#6}{ and~\S\S#3#1#4--#5#2#6}
\Crefrangemultiformat{section}{\S\S#3#1#4--#5#2#6}{ and~\S\S#3#1#4--#5#2#6}{, \S\S#3#1#4--#5#2#6}{ and~\S\S#3#1#4--#5#2#6}

\usepackage[normalem]{ulem}
\usepackage{contour}

\contourlength{0.8pt}
\newcommand{\myuline}[1]{%
  \uline{\phantom{#1}}%
  \llap{\contour{white}{#1}}%
}

\usepackage[T1]{fontenc}

\usepackage[utf8]{inputenc}

\usepackage{microtype}
\usepackage{xspace}

\newcommand{\lm}{\mathrm{LM}}

\renewcommand{\vec}[1]{\mathbf{#1}}
\newcommand{\pP}{\vec{t}}

\newcommand{\PP}{\mathcal{T}}
\newcommand{\px}{x}
\newcommand{\py}{y}

\newcommand{\pr}{r}

\newcommand{\db}{\mathcal{E}}

\newcommand{\reimp}{$^\dagger$}
\newcommand{\notcomparable}{$^?$}

\newcommand{\blank}{\underline{\hspace{4ex}}\xspace}
\newcommand{\blankx}{\blank$\!_\mathrm{x}$\xspace}
\newcommand{\blanky}{\blank$\!_\mathrm{y}$\xspace}

\usepackage{todonotes}
\makeatletter
\newcommand*\iftodonotes{\if@todonotes@disabled\expandafter\@secondoftwo\else\expandafter\@firstoftwo\fi}  %
\makeatother

\newlength{\extramargin}
\usepackage{calc}
\iftodonotes{\geometry{paperwidth=\paperwidth+\extramargin*4,marginparwidth=\marginparwidth+\extramargin,width=\textwidth,height=\textheight}}

\usepackage{ctable}

\title{Learning How to Ask: Querying LMs with Mixtures of Soft Prompts}

\author{Guanghui Qin \and Jason Eisner\\ Department of Computer Science, Johns Hopkins University 
\\ {\tt gqin2@jhu.edu \quad jason@cs.jhu.edu} \\}

\begin{document}
\maketitle
\begin{abstract}
Natural-language prompts have recently been used to coax pretrained language models into performing other AI tasks, using a fill-in-the-blank paradigm \citep{petroniLanguageModelsKnowledge2019} or a few-shot extrapolation paradigm \cite{brownLanguageModelsAre2020}.  For example, language models retain factual knowledge from their training corpora that can be extracted by asking them to ``fill in the blank'' in a sentential prompt.  However, where does this prompt come from?  We explore the idea of learning prompts by gradient descent---either fine-tuning prompts taken from previous work, or starting from random initialization.  Our prompts consist of ``soft words,'' i.e., continuous vectors that are not necessarily word type embeddings from the language model.  Furthermore, for each task, we optimize a mixture of prompts, learning which prompts are most effective and how to ensemble them.  
Across multiple English LMs and tasks, our approach hugely outperforms previous methods, showing that the implicit factual knowledge in language models was previously underestimated.  Moreover, this knowledge is cheap to elicit: random initialization is nearly as good as informed initialization.
\end{abstract}

\section{Introduction}\label{sec:intro}

Pretrained language models, such as ELMo \citep{petersDeepContextualizedWord2018}, BERT \citep{devlinBERTPretrainingDeep2019}, and BART \citep{lewisBARTDenoisingSequencetoSequence2020}, have proved to provide useful representations for other NLP tasks.  Recently, \citet{petroniLanguageModelsKnowledge2019} and \citet{jiangHowCanWe2020} demonstrated that language models (LMs) also contain factual and commonsense knowledge that can be elicited with a prompt.
For example, to query the \texttt{date-of-birth} of Mozart, we can use the prompt ``\myuline{Mozart} was born in \blank,'' where we have filled the first blank with ``Mozart,'' and ask a cloze language model to fill in the second blank.
The prompts used by \citet{petroniLanguageModelsKnowledge2019} are manually created, while \citet{jiangHowCanWe2020} use mining and paraphrasing based methods to automatically augment the prompt sets.

Finding out what young children know is difficult because they can be very sensitive to the form of the question \cite{donaldsonChildrenMinds1978}.  Opinion polling is also sensitive to question design \citep{broughton1995assumptions}.
We observe that when we are querying an LM rather than a human, we have the opportunity to \emph{tune} prompts using gradient descent---the workhorse of modern NLP---so that they better elicit the desired type of knowledge.  

A neural LM sees the prompt as a sequence of continuous word vectors \citep{baroniDonCountPredict2014}.  We tune in this continuous space, relaxing the constraint that the vectors be the embeddings of actual English words.  Allowing ``soft prompts'' consisting of ``soft words'' is not only convenient for optimization, but is also more expressive.  Soft prompts can emphasize particular words (by lengthening their vectors) or particular dimensions of those words.  They can also adjust words that are misleading, ambiguous, or overly specific.  Consider the following prompt for the relation \texttt{date-of-death}:
\begin{quote}
\mbox{\blankx performed until his death in \blanky.}
\end{quote}
This prompt may work for the male singer Cab Calloway, but if we want it to also work for the female painter Mary Cassatt, it might help to soften ``performed'' and ``his'' so that they do not insist on the wrong occupation and gender, and perhaps to soften ``until'' into a weaker connective (as Cassatt was in fact too blind to paint in her final years).

Another way to bridge between these cases is to have one prompt using ``performed'' and another using ``painted.''  In general, there may be many varied lexical patterns that signal a particular relation, and having more patterns will get better coverage
\cite{hearstAutomaticAcquisitionHyponyms1992,riloffLearningDictionariesInformation1999}.  We therefore propose to learn a \emph{mixture} of soft prompts.

We test the idea on several cloze language models, training prompts to complete factual and common sense relations from 3 datasets.  Comparing on held-out examples, our method dramatically outperforms previous work, even when initialized randomly.  So when regarded as approximate knowledge bases, language models know more than we realized.  We just had to find the right ways to ask.

\section{Related Work}

Factual knowledge is traditionally extracted from large corpora using a pipeline of NLP tools \citep{surdeanuOverviewEnglishSlot2014}, including entity extraction \citep{lampleNeuralArchitecturesNamed2016}, entity linking \citep{rao2013entity} and relation extraction \citep{sorokinContextAwareRepresentationsKnowledge2017}.

However, recent work has shown that simply training a system to complete sentences---language modeling---causes it to implicitly acquire non-linguistic abilities from its training corpora \citep{rogersPrimerBERTologyWhat2020}, including factual knowledge \citep{petroniLanguageModelsKnowledge2019,jiangHowCanWe2020}, common sense \citep{biskPIQAReasoningPhysical2019}, reasoning \citep{talmorOLMpicsWhatLanguage2020,brownLanguageModelsAre2020},
summarization \citep{radfordLanguageModelsAre2019}, 
and even arithmetic \citep{bouraouiInducingRelationalKnowledge2020}.

Most of the previous work manually creates prompts to extract answers from the trained language model.  We use LAMA \cite{petroniLanguageModelsKnowledge2019} as a baseline.  Building on LAMA, the LM Prompt And Query Archive (LPAQA) method \cite{jiangHowCanWe2020} searches for new prompts by either mining a corpus or paraphrasing existing prompts.  AutoPrompt \cite{shinAutoPromptElicitingKnowledge2020} searches for improved prompts using a gradient signal, although its prompts are limited to sequences of actual (``hard'') English words, unlike our method.  We compare our novel soft prompts against all of these systems.

After we submitted the present paper in November 2020, two still unpublished manuscripts appeared on arXiv that also investigated soft prompts.  \Citet{li2021prefix} considered the setting of generating text from a pretrained language model (GPT-2 or BART) conditioned on a textual prompt.  To improve the results, they prepended a few task-specific ``soft tokens'' to the prompt and tuned the embeddings of only these tokens (at all embedding layers).  \citet{liu2021gpt} adopted a strategy similar to ours by tuning fill-in-the-blank prompts in a continuous space, testing on GPT-2 and BERT models, although they did not use the enhancements we proposed in \crefrange{sec:deep-soft-prompts}{sec:data-dep-weights} below.  Like our work, both these papers achieved strong gains.%

In other work, \citet{bouraouiInducingRelationalKnowledge2020} mine prompts from a corpus, then fine-tune the whole language model so that it more accurately completes the prompts.  \citet{schick2020exploiting,schick2020s} are similar but fine-tune the language model differently for each prompt.  Our method complements these by tuning the prompts themselves.

``Probing'' systems that ask what language models know \emph{about particular sentences} \cite[e.g.,][]{eichler-etal-2019-linspector} usually use feedforward networks rather than further natural-language prompts.  Yet \citet{shinAutoPromptElicitingKnowledge2020} show how to use natural-language prompts to ask about particular sentences.  Our method could potentially be applied to those prompts, or to ``few-shot learning'' prompts that include input-output examples \cite{brownLanguageModelsAre2020}.

\section{Method}

Our experiments will specifically aim at extracting relational knowledge from language models.  We are given a fixed pretrained LM, a specific binary relation $\pr$ such as \texttt{date-of-death}, and a training dataset $\db_r$ consisting of known $(\px,\py)$ pairs in $\pr$, such as (Mary Cassatt, 1926).  We will then train a system to predict $\py$ from $\px$, and evaluate it on held-out $(\px,\py)$ pairs of the same relation.

A prompt $\pP$ is a sentence or phrase that includes two blanks, as illustrated in \cref{sec:intro}. To pose the query, we fill the \blankx blank with $\px$:
\begin{quote}
\myuline{Mary Cassatt} performed until his death in \blanky.
\end{quote}
We can ask the LM for its probability distribution $p_\lm(y \mid \pP, \px)$ over single words that can now fill \blanky.  The correct answer would be 1926.

\subsection{Soft Prompts}\label{sec:soft-prompts}

Suppose the LM identifies the word types with vectors in $\mathbb{R}^d$.
We also allow $\pP$ to be a soft prompt, in which the tokens can be arbitrary vectors in $\mathbb{R}^d$:
\begin{quote}
\mbox{\blankx $\;v_1$ $\;v_2$ $\;v_3$ $\;v_4$ $\;v_5$ $\;$\blanky $\;v_6$}
\end{quote}
We can initialize these vectors to match those of a given hard prompt.  (Each token of a hard prompt may be a word, subword, or punctuation mark, according to the tokenization procedure used by the LM.)  However, we can then tune the vectors continuously.  We do not change the number of vectors or their positions.  For the prompt shown above, we have a $6d$-dimensional search space.

\subsection{Deeply Perturbed Prompts}\label{sec:deep-soft-prompts}

For each token $i$ of a prompt, the vector $v_i$ enters into the LM's computations that complete the prompt.  For example, a Transformer architecture computes successively deeper contextual embeddings of the token, $v_i^{(\ell)}: 0 \leq \ell \leq L$.  Here $v_i^{(0)} = v_i$ and the embedding $v_i^{(\ell)}$ at layer $\ell > 0$ is computed from all tokens' embeddings $v_j^{(\ell-1)}$ at the previous layer, using the LM's parameters.

We can tune the prompt by additively perturbing each $v_i^{(\ell)}$ by a small vector $\Delta_i^{(\ell)}$ before it is used in further computations.  The $\Delta$ vectors for a given hard prompt are initialized to 0 and then tuned. 

Perturbing only layer 0 is equivalent to tuning $v_i$ directly as in \cref{sec:soft-prompts}.  However, if we are more aggressive and perturb all layers, we now have $6d\cdot(L+1)$ parameters to tune a 6-token prompt.  The perturbations ($\Delta$ vectors) can be kept small through early stopping or some other form of regularization.
Our intuition is that small perturbations will yield more ``familiar'' activation patterns that are similar to those that the LM was originally trained on.  (\citet{li2021prefix} tried a rather different approach to preventing overfitting when tuning all layers.)

\subsection{Mixture Modeling}\label{sec:mixture}

Given a set $\PP_r$ of soft prompts for relation $r$, we can define the ensemble predictive distribution
\begin{align}
    p(\py \mid \px, \pr) = \sum_{\pP \in \PP_r} p(\pP \mid \pr) \cdot p_\lm(y \mid \pP, \px) \label{eq:mix_prob}
\end{align}
where the learned mixture weights $p(\pP \mid \pr)$ form a distribution over the soft prompts $\pP \in \PP_r$.  
Ensembling techniques other than mixture-of-experts could also be
used, including product-of-experts \cite{jiangHowCanWe2020}.

\subsection{Data-Dependent Mixture Modeling}\label{sec:data-dep-weights}

As an extension, we can replace the mixture weights $p(\pP \mid \pr)$ with $p(\pP \mid \pr, \px)$, to allow the model to select prompts that are appropriate for the given $\px$.  For example, a plural noun $\px$ might prefer prompts $\pP$ that use a plural verb.

While we could directly build a neural softmax model for $p(\pP \mid \pr, \px)$, it seems useful to capture the intuition that $\pP$ may work better if $\px$ is plausible in its \blankx.  Thus, we instead use Bayes' Theorem to write $p(\pP \mid \pr, \px)$ as proportional to $p(\pP \mid \pr) \cdot p(\px \mid \pP, \pr)^{1/T}$, where we have included $T$ to modulate the strength of the above intuition.\footnote{Raising the temperature $T$ increases the entropy of the mixture to get the benefits of ensembling; without $T$, the strong language model usually places almost all the weight on a single prompt.}  Here $p(\pP \mid \pr)$ is still a learned distribution over prompts, and we use the fixed language model to estimate the second factor as $\sum_{\py} p_\lm(\px, \py \mid \pP)$ (dropping the dependence on $\pr$ just as we did for the second factor of \eqref{eq:mix_prob}).  $\log T$ is tuned along with all other parameters.

\subsection{Training Objective}
\label{sec:trainobj}

Given an initial set of prompts $\PP_\pr$, we jointly optimize the soft prompts $\pP \in \PP$ and their mixture weights $p(\pP \mid \pr)$ (and $\log T$ in \cref{sec:data-dep-weights}) to minimize the log-loss of the predictive distribution \eqref{eq:mix_prob}:\looseness=-1
\begin{align}
    \sum_{(x,y) \in \db_\pr} -\log \sum_{\pP \in \PP_\pr} p(\py \mid \pP, \px)
    \label{eq:objective}
\end{align}

This is a continuous and differentiable objective whose gradient can be computed by back-propagation.  It can be locally minimized by gradient descent (using a softmax parameterization of the mixture weights).
Equivalently, it can be locally minimized by the EM algorithm: the E step finds a posterior distribution over latent prompts for each $(x,y)$ example, and the M step performs gradient descent to optimize the prompts in that mixture.

\section{Experiments}
\setlength{\textfloatsep}{11pt plus 6.0pt minus 2.0pt}     %

\subsection{Relational Datasets} 

The relations we learn to predict are T-REx original \citep{elsaharTRExLargeScale2018},  T-REx extended \citep{shinAutoPromptElicitingKnowledge2020}, Google-RE \citep{google-re}, and ConceptNet \citep{speerConceptNetOpenMultilingual2017}---or rather, the subsets 
that were used by the LAMA and AutoPrompt papers.
See \cref{app:rel_stats} for some statistics.

\subsection{Language Models} 

Following \citet{petroniLanguageModelsKnowledge2019}, we interrogate BERT  \citep{devlinBERTPretrainingDeep2019} and RoBERTa  \citep{liuRoBERTaRobustlyOptimized2019}. 
These are masked (cloze) language models.  
For variety, we also interrogate BART \citep{lewisBARTDenoisingSequencetoSequence2020}, which conditions on the prompt with empty \blanky and generates a copy where \blanky has been filled in (by a single token).  We constrain BART's decoding to ensure that its answer does take this form.  Unlike BERT and RoBERTa, BART could be used to fill \blanky with an arbitrarily long phrase, but we do not allow this because $y$ in our datasets is always a single token.\footnote{Among other filters, the LAMA and AutoPrompt papers keep only the triples $(r,x,y)$ such that $y$ is a single token according to the language models used by LAMA.  When working with BART, we further require $y$ to be a single token according to BART's tokenization; thus, the BART results are not comparable with the other language models.}

\subsection{Dataset Splits}\label{sec:splits}

For the two T-REx datasets, we inherit the training-validation-test split from \citet{shinAutoPromptElicitingKnowledge2020}.  For the other datasets, we split randomly in the ratio 80-10-10.\footnote{The LAMA paper \citep{petroniLanguageModelsKnowledge2019} provided no split but used everything as test data for their zero-shot method.}  Since all pairs $(\px,\py)$ are distinct, there are no common triples among these three sets.  Common $\px$ values are also rare because each dataset has at least 174 distinct $x$ values.  However, the number of distinct $y$ values can be as small as 6.  Thus, in another set of experiments (\cref{sec:challenging}), we used a more challenging split that ensures that there are no common $\py$ values among these three sets.  This tests whether our model generalizes to unseen values.\looseness=-1

\subsection{Prompts} 

For the T-REx and Google-RE datasets, we have four sources of initial prompts:
\begin{itemize}
    \item (sin.) LAMA provides a \textbf{sin}gle manually created hard prompt for each relation type $r$.
    \item (par.) LPAQA \citep{jiangHowCanWe2020} provides a set of 13--30 hard prompts for each $r$, which are \textbf{par}aphrases of the LAMA prompt.\footnote{The LPAQA system combines their predictions via a learned weighted product of experts.}
    \item (min.) LPAQA also provides a set of 6--29 hard prompts for each $r$, based on text \textbf{min}ing.
    \item (ran.) For each (min.) prompt, we replace each word with a \textbf{ran}dom vector, drawn from a Gaussian distribution fit to all of the LM's word embeddings.  The number of words and the position of the blanks are preserved.
\end{itemize}

For the ConceptNet dataset, LAMA uses the gold Open Mind Common Sense (OMCS) dataset \citep{singhOpenMindCommon2002}.  In this dataset, each example $(\px_i,\py_i)$ is equipped with its own prompt $\pP_i$.  (Each example is really a sentence with two substrings marked as $\px$ and $\py$, which are removed to obtain $\pP_i$.)  These prompts are often overly specific: often $\py_i$ can be predicted from $(\pP_i,\px_i)$, or just from $\pP_i$ alone, but $\py_j$ cannot be predicted from $(\pP_i,\px_j)$.  Thus, for each relation $\pr$, we use only the prompts that appear more than 10 times, resulting in 1--38 prompts.    

Statistics about the prompts are in \cref{app:prompt_stats}.\looseness=-1

We used only a single copy of each prompt, but a generalization would be to allow multiple slightly perturbed copies of each prompt, which could diverge and specialize during training \cite{rose-1998}.

\subsection{Training}

We optimize \cref{eq:objective} with the method introduced in \cref{sec:trainobj}.
We use the Adam optimizer \citep{kingmaAdamMethodStochastic2015} with its default configuration.
For gradient training, we set the batch size as 64, early-stop patience as 4, and test with the model that performs best on the dev set among 16 training epochs.

Training is fast.  Even for our largest model (BERT-large-cased) and largest dataset (T-REx extended), tuning a single prompt completes within a few minutes.  With a mixture of prompts, training scales roughly linearly with the number of prompts.  It is still presumably much cheaper in time and memory than fine-tuning the entire BERT model, which must back-propagate a much larger set of gradients.%

\subsection{Metrics and Baselines}

Our method outputs the most probable $\py$ given $(\pr,\px)$.  Here and in the supplementary material, we report its average performance on all test examples,
with precision-at-1 (P@1), precision-at-10 (P@10) and mean reciprocal rank (MRR) as metrics.  We measure the improvement from tuning LAMA, LPAQA, and random prompts.  We also compare with AutoPrompt. 
Baseline numbers come from prior papers or our reimplementations.

\subsection{Results}

\begin{table}[t]
\centering
\begin{tabular}{@{\hspace{0.0\tabcolsep}}c@{\hspace{0.7\tabcolsep}}|@{\hspace{0.7\tabcolsep}}l@{\hspace{0.5\tabcolsep}}l@{\hspace{0.\tabcolsep}}}
\specialrule{.1em}{.05em}{.05em}
Model & \multicolumn{1}{c}{T-REx orig.}  & \multicolumn{1}{c}{T-REx ext.} \\
\specialrule{.1em}{.05em}{.05em}
 \textsc{lama} (BEb) & 31.1  & 26.4\\
 \textsc{lpaqa}(BEb)  & 34.1  & 31.2  \\
 AutoPrompt & 43.3  & 45.6  \\
 Soft (sin., BEb) & 47.7 {\small (+\textbf{16.6}\notcomparable) } & 49.6 {\small (+\textbf{23.2}\notcomparable) }\\
 Soft (min., BEb) & \textbf{50.7}\notcomparable {\small (+\textbf{16.6}\notcomparable) } & \textbf{50.5}\notcomparable {\small (+\textbf{19.3}\notcomparable) } \\
 Soft (par., BEb) & 48.4 {\small (+\textbf{12.8}\notcomparable) } & 49.7 {\small (+\textbf{18.5}\notcomparable) } \\
 Soft (ran., BEb) & 48.1 {\small (+\textbf{47.4}) } & 50.6 {\small (+\textbf{49.8}) } \\
\hline
 \textsc{lama} (BEl) & 28.9\reimp & 24.0\reimp \\
 \textsc{lpaqa}(BEl) & 39.4\reimp  & 37.8\reimp  \\
 Soft (sin., BEl) & 51.1 {\small (+\textbf{22.2}) } &  51.4 {\small (+\textbf{27.4}) } \\
 Soft (min., BEl) & \textbf{51.6} {\small (+\textbf{12.2}) } & \textbf{52.5} {\small (+\textbf{14.7}) }  \\
 Soft (par., BEl) & 51.1 {\small (+\textbf{11.7}) } & 51.7 {\small (+\textbf{13.9}) } \\
 Soft (ran., BEl) & 51.9 {\small (+\textbf{47.1}) } & 51.9 {\small (+\textbf{50.5}) } \\
 \hline
 AutoPrompt & 40.0 & -  \\
 Soft (min., Rob) & \textbf{40.6}\notcomparable {\small (+\textbf{39.4}) } & - \\
\specialrule{.1em}{.05em}{.05em}
\end{tabular}
\caption{\label{tab:trex}
Results on T-REx datasets with P@1 as the metric.
The ``Soft'' lines (our method) parenthetically show the improvement over 
the initial parameters (boldfaced if significant).
In each subcolumn of comparable results, we boldface the best
result along with all that are not significantly worse (sign test, $p < 0.02$).  
(We marked a boldface number with "?" if we lacked access to per-example output for one of the systems; differences from such systems were simply assumed to be significant.)
\reimp{} marks baseline results obtained from our reimplementations. 
In the Model column, BEb is BERT-base, BEl is BERT-large, Rob is RoBERTa-base.}
\end{table}

\Cref{tab:trex} shows results on T-REx datasets obtained by querying three BERT-style models, with P@1 as the metric.  Additional metrics and language models are shown in \cref{tab:google-re,tab:conceptnet} as well as \cref{tab:trex-ori-full,tab:trex-ext-full} in the supplementary material.  

We consistently get large improvements by tuning the initial prompts.  Remarkably, our method beats all prior methods even when throwing away the words of their informed prompts in favor of random initial vectors.  It simply finds a prompt that works well on the $(x,y)$ training examples.

\begin{table}[t]
\centering
\begin{tabular}{@{\hspace{0.0\tabcolsep}}c@{\hspace{0.5\tabcolsep}}|@{\hspace{0.5\tabcolsep}}l@{\hspace{0.5\tabcolsep}}l@{\hspace{0.5\tabcolsep}}l@{\hspace{0.0\tabcolsep}}}
\specialrule{.1em}{.05em}{.05em}
Model & \multicolumn{1}{c}{P@1} & \multicolumn{1}{c}{P@10} & \multicolumn{1}{c}{MRR}\\
\specialrule{.1em}{.05em}{.05em}
\textsc{lama} & \ \;9.7\reimp & 27.0\reimp & 15.6\reimp \\
\textsc{lpaqa} & 10.6\reimp & 23.7\reimp & 15.3\reimp \\
\hline
Soft (sin.) & \textbf{11.2} {\small (+\textbf{1.5})} &\textbf{33.5} {\small (+\textbf{\ \;6.5})} & \textbf{18.9} {\small (+\textbf{3.3})} \\
Soft (min.) & \textbf{12.9} {\small (+\textbf{2.3})} & \textbf{34.7} {\small (+\textbf{11.0})} & \textbf{20.3} {\small (+\textbf{5.0})} \\
Soft (par.) & \textbf{11.5} {\small (+\textbf{0.9})} & \textbf{31.4} {\small (+\ \;\textbf{7.7})} & \textbf{18.3} {\small (+\textbf{3.0})} \\
\specialrule{.1em}{.05em}{.05em}
\end{tabular}
\caption{\label{tab:google-re}
Results on Google-RE dataset obtained by querying the BERT-large-cased model.}
\end{table}

\begin{table}[t]
\centering
\begin{tabular}{@{\hspace{0.0\tabcolsep}}c@{\hspace{0.2\tabcolsep}}|@{\hspace{0.2\tabcolsep}}l@{\hspace{0.5\tabcolsep}}l@{\hspace{0.5\tabcolsep}}l@{\hspace{0.0\tabcolsep}}}
\specialrule{.1em}{.05em}{.05em}
Model & \multicolumn{1}{c}{P@1} & \multicolumn{1}{c}{P@10} & \multicolumn{1}{c}{MRR}\\
\specialrule{.1em}{.05em}{.05em}
\textsc{lama} (BEb) & \ \;0.1\reimp & \ \;2.6\reimp & \ \;1.5\reimp \\
\textsc{lama} (BEl) & \ \;0.1\reimp & \ \;5.0\reimp & \ \;1.9\reimp \\
\hline
Soft (min.,BEb) & 11.3{\small(+\textbf{11.2})} & 36.4{\small(+\textbf{33.8})} & 19.3{\small(+\textbf{17.8})} \\
Soft (ran.,BEb) & \textbf{11.8}{\small(+\textbf{11.8})} & \textbf{34.8}{\small(+\textbf{31.9})} & \textbf{19.8}{\small(+\textbf{19.6})} \\
Soft (min.,BEl) & \textbf{12.8}{\small(+\textbf{12.7})} & \textbf{37.0}{\small(+\textbf{32.0})} & \textbf{20.9}{\small(+\textbf{19.0})} \\
Soft (ran.,BEl) & \textbf{14.5}{\small(+\textbf{14.5})} & \textbf{38.6}{\small(+\textbf{34.2})} & \textbf{22.1}{\small(+\textbf{21.9})} \\
\specialrule{.1em}{.05em}{.05em}
\end{tabular}
\caption{
\label{tab:conceptnet}
Results on ConceptNet (winner: random init).
}
\end{table}

\begin{table}[!t]
\centering
\begin{tabular}{c|ccc}
\specialrule{.1em}{.05em}{.05em}
Model & P@1 & P@10 & MRR\\
\specialrule{.1em}{.05em}{.05em}
baseline & 39.4 & 67.4 & 49.1 \\
$\!\!$adjust mixture weights & 40.0 & 69.1 & 53.3 \\
adjust token vectors & 50.7 & 80.7 & 61.1 \\
adjust both & \textbf{51.0} & \textbf{81.4} & \textbf{61.6} \\
\specialrule{.1em}{.05em}{.05em}
\end{tabular}
\caption{\label{tab:ablation}
Ablation experiments, conducted with the BERT-large model on the T-REx original dataset.}
\end{table}

We conduct an ablation study where we adjust only the mixture weights 
(which are initially uniform)
or only the word vectors in the prompts $\pP$. As \cref{tab:ablation} shows, each helps, but the major benefit comes from tuning the word vectors to get soft prompts.  \Cref{app:vis} visualizes a set of soft prompts, and \cref{app:eff_prompts} analyzes the mixture weights.
We also experiment on a challenging setting where the $\py$ labels are distinct for training and test (\cref{sec:challenging} in the supplementary materials),
and find that soft prompts still yield some benefits.

The above results are for our basic method that tunes only the words of the prompt (i.e., layer 0).  When we tune all layers---the ``deeply perturbed prompts'' of \cref{sec:deep-soft-prompts}---we typically obtain small additional gains, across various models and initializations, although tuning all layers does substantially hurt RoBERTa.
These results are shown in \cref{tab:trex-ori-full,tab:trex-ext-full} in the supplementary material.

The tables show that the winning system---for \emph{each} combination of language model, T-REx dataset, and evaluation metric---\emph{always} uses a mixture of soft prompts initialized to mined prompts.  It always tunes all layers, except with RoBERTa.

Finally, we also tried using data-dependent mixture weights as in \cref{sec:data-dep-weights}.  This had little effect, because training learned to discard the $x$ information by setting the temperature parameter $T$ high.

\section{Conclusion}

Well-crafted natural language prompts are a powerful way to extract information
from pretrained language models.  In the case of cloze prompts used
to query BERT and BART models for single-word answers, we have demonstrated 
startlingly large and consistent improvements from rapidly learning 
prompts that work---even though the resulting ``soft prompts'' 
are no longer natural language.  

Our code and data are available at \url{https://github.com/hiaoxui/soft-prompts}.

How about few-shot prediction with pretrained generative LMs? Here, \citet{lewis2020etrievalaugmented} show how to assemble a natural language prompt for input $\px$ from relevant input-output pairs $(\px_i,\py_i)$ selected by a trained retrieval model.  Allowing fine-tuned \emph{soft} string pairs is an intriguing future possibility for improving such methods without needing to fine-tune the entire language model.

\clearpage

\section*{Acknowledgments}

We thank the anonymous reviewers for helpful comments.
This work was supported by DARPA KAIROS and by the National Science Foundation under Grant No.\@ 1718846. The U.S. Government is authorized to reproduce and distribute reprints for governmental purposes. The views and conclusions contained in this publication are those of the authors, and should not be interpreted as representing official policies nor endorsement by the funding agencies or by Microsoft (where Dr.\@ Eisner is also a paid employee, in an arrangement that has been reviewed and approved by the Johns Hopkins University in accordance with its conflict of interest policies).\looseness=-1

\bibliography{ref,custom}
\bibliographystyle{acl_natbib}

\clearpage

\appendix

\section{Statistics of Relational Databases}\label{app:rel_stats}

The statistics of the various relational databases are shown in \cref{tab:rel_stats}.

\section{Statistics of the Initial Prompts}\label{app:prompt_stats}

\Cref{tab:prompt_stats} shows some statistics of the prompts we use to initialize the SoftPrompt model.

\section{Visualization of Soft Prompts}\label{app:vis}

\begin{figure}[h]
    \centering
    \includegraphics[scale=0.6]{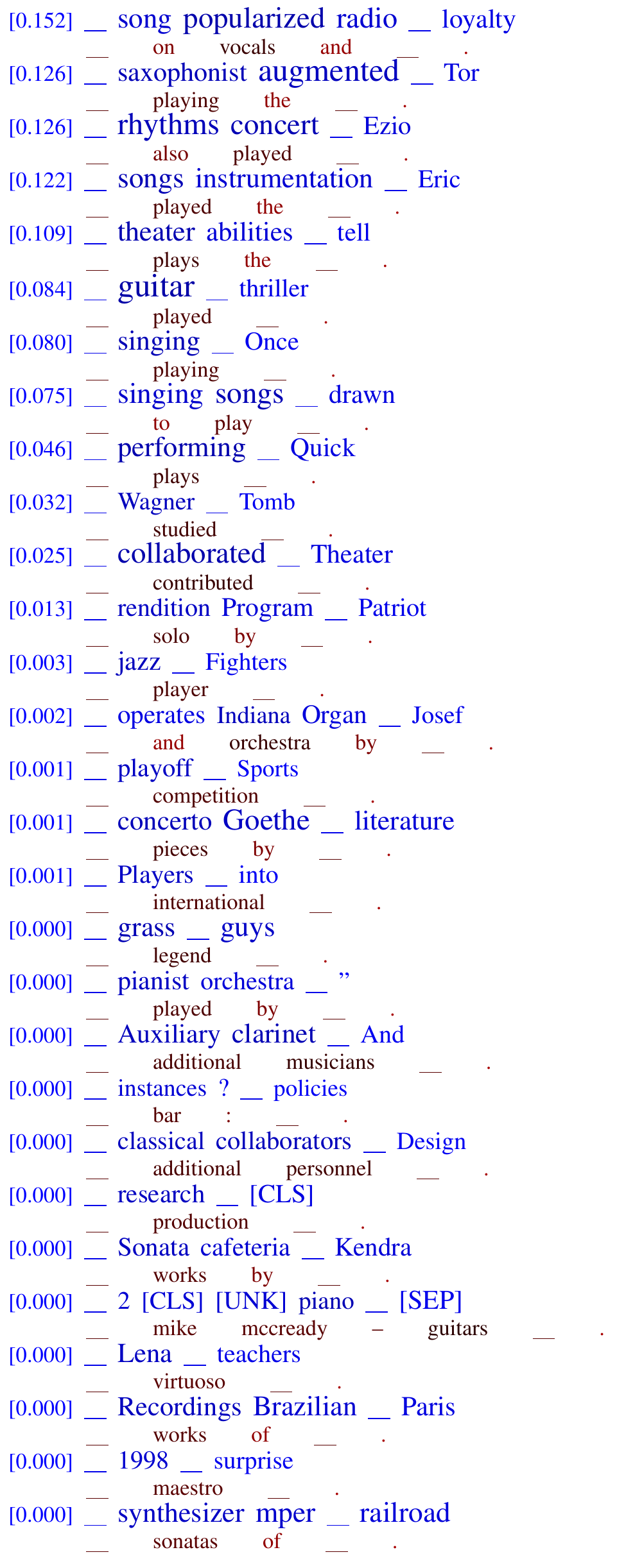}
    \caption{
    Visualization of the LPAQA mining prompts for relation \texttt{P1303 Instrument} (i.e., $\px$ plays instrument $\py$) from T-REx extended. We show the effect of tuning the layer-0 token embeddings (but not higher layers) on BERT-large-cased.  The prompts are sorted in decreasing order by mixture weight.  Each prompt's weight is shown at left; note that after the first 12 prompts, the remaining ones have negligible contribution.  We show each soft prompt in blue, followed by the original (mined) prompt in red.
    To visualize the tuned vector $\vec{v}$, we display the blue word $w$ that maximizes $p(w \mid \vec{v})$.  The brightness of the blue word $w$ and the original red word $w_0$ are respectively proportional to $p(w \mid \vec{v})$ and $p(w_0 \mid \vec{v})$.  The red word has size 1, and the blue word has size $||\vec{v}|| / ||\vec{v}_0||$, where $\vec{v}_0$ is the
    original untuned vector (the embedding of $w_0$).
    In this example, the
    blue probabilities $p(w \mid \vec{v})$
    range from 6.5e-5 to 9.7e-5 (mean 8.6e-5 $\pm$ 8.1e-6), the
    red probabilities $p(w_0 \mid \vec{v})$ range from 7.7e-5 to 1.1e-4 (mean 9.5e-5 $\pm$ 7.8e-6), and the relative magnitudes $||\vec{v}|| / ||\vec{v}_0||$ vary from 1.00 to 1.49 (mean 1.12 $\pm$ 0.13).
    }
    \label{fig:case}
\end{figure}

\Cref{fig:case} shows what a mixture of soft prompts looks like when we tune only layer 0.  The soft prompts are not too interpretable.  The words closest to the tuned tokens (shown in blue) seem to be largely on the music topic.  However, the soft templates do not seem to form meaningful phrases, nor is it obvious why they would prime for $y$ to be an instrument when $x$ is a musician.

\section{Entropy of the Mixture Model}\label{app:eff_prompts}

For any given relation $r$, the entropy of the mixture weights is
\begin{align}
    H &= \sum_{\pP \in \PP_r} p(\pP \mid \pr) \cdot \log_2 p(\pP \mid \pr)
\end{align}
We then take $2^H \in [1, |\PP_r|]$ as a measure of the effective number of prompts that were retained. 
\Cref{tab:eff_prompts} shows some statistics of the effective number of prompts.
In some cases, tuning the mixture weights essentially selected a single prompt, but on average, it settled on a mixture of several variant prompts (as illustrated by \cref{fig:case}).

\section{Challenging dataset with distinct $y$'s}
\label{sec:challenging}

As described in \cref{sec:splits}, we conducted an additional experiment to determine whether the prompts could generalize to novel $y$ values.
We conduct another experiment and ensure that there are no common $\py$ values among the train / dev / test sets.
We use T-REx as the base relational database and split the datasets to make the ratio close to 80-10-10.
The experiment results are shown in \cref{tab:distinct}.
We can observe that our method again improves the results, just as in \cref{tab:trex-ori-full,tab:trex-ext-full},
which shows the generalizability of our method.

\begin{table*}[b]
\centering
\begin{tabular}{@{\hspace{0.2\tabcolsep}}c@{\hspace{0.2\tabcolsep}}|@{\hspace{0.1\tabcolsep}}c@{\hspace{0.1\tabcolsep}}@{\hspace{0.2\tabcolsep}}|@{\hspace{0.2\tabcolsep}}l@{\hspace{0.2\tabcolsep}}|@{\hspace{0.2\tabcolsep}}l@{\hspace{0.2\tabcolsep}}|@{\hspace{0.2\tabcolsep}}l}
\specialrule{.1em}{.05em}{.05em}
\multirow{2}{*}{LM} & \multirow{2}{*}{Method} & \multicolumn{1}{c}{Precision@1} & \multicolumn{1}{c}{Precision@10} & \multicolumn{1}{c}{MRR}\\
\cline{3-5}
 &  & \ init\ \ \ \,$\rightarrow$\ \ \ \,soft\ \ \ \,$\rightarrow$\ \ \ \,deep & \ init\ \ \ \,$\rightarrow$\ \ \ \,soft\ \ \ \,$\rightarrow$\ \ \ \,deep & \ init\ \ \ \,$\rightarrow$\ \ \ \,soft\ \ \ \,$\rightarrow$\ \ \ \,deep\\
 \hline
\multirow{6}{*}{BEb} & \textsc{lama} & 31.1 & 59.5 & 40.3\\
 & \textsc{lpaqa} & 34.1 & 62.0 & 43.6\\
 & Soft (sin.) & 31.1 {\tiny$\xrightarrow{  \mathbf{ +14.6^? }  } $} 45.7 {\tiny$\xrightarrow{  \mathbf{ +\ \ 2.0 }  }$} 47.7 & 59.5 {\tiny$\xrightarrow{  \mathbf{ +16.3^? }  }$} 75.8 {\tiny$\xrightarrow{  \mathbf{ +\ \ 3.2 }  }$} 79.0 & 40.3 {\tiny$\xrightarrow{  \mathbf{ +15.9^? }  }$} 56.2 {\tiny$\xrightarrow{  \mathbf{ +\ \ 2.2 }  }$} 58.4\\
 & Soft (min.) & 34.1 {\tiny$\xrightarrow{  \mathbf{ +14.7^? }  }$} 48.8 {\tiny$\xrightarrow{  \mathbf{+\ \ 1.9 }  }$} \textbf{50.7}\notcomparable & 62.0 {\tiny$\xrightarrow{  \mathbf{ +15.6^? }  }$} 79.6 {\tiny$\xrightarrow{  \mathbf{ +\ \ 1.1 }  }$} \textbf{80.7}\notcomparable & 43.6 {\tiny$\xrightarrow{  \mathbf{ +15.8^? }  }$} 59.4 {\tiny$\xrightarrow{  \mathbf{ +\ \ 1.7 }  }$} \textbf{61.1}\notcomparable\\
 & Soft (par.) & 34.1 {\tiny$\xrightarrow{  \mathbf{ +12.8^? }  }$} 46.9 {\tiny$\xrightarrow{  \mathbf{ +\ \ 1.5 }  }$} 48.4 & 62.0 {\tiny$\xrightarrow{  \mathbf{ +16.8^? }  }$} 78.8 {\tiny$\xrightarrow{  \mathbf{ +\ \ 0.8 }  }$} 79.6 & 43.6 {\tiny$\xrightarrow{  \mathbf{ +14.2^? }  }$} 57.8 {\tiny$\xrightarrow{  \mathbf{ +\ \ 1.3 }  }$} 59.1\\
 & Soft (ran.) & \ \ 0.7 {\tiny$\xrightarrow{  \mathbf{+46.6\ \ }  }$} 47.3 {\tiny$\xrightarrow{  \mathbf{ +\ \ 0.8 }  }$} 48.1 & \ \ 4.6 {\tiny$\xrightarrow{  \mathbf{+74.0\ \ }  }$} 79.1 {\tiny$\xrightarrow{  +\ \ 0.0 }$} 79.1 & \ \ 2.3 {\tiny$\xrightarrow{  \mathbf{+56.1\ \  }  }$} 58.4 {\tiny$\xrightarrow{  \mathbf{ +\ \ 0.5 }  }$} 58.9\\
 \hline
\multirow{6}{*}{BEl} & \textsc{lama} & 28.9\reimp & 57.7\reimp & 38.7\reimp\\
 & \textsc{lpaqa} & 39.4\reimp & 67.4\reimp & 49.1\reimp\\
 & Soft (sin.) & 28.9 {\tiny$\xrightarrow{  \mathbf{ +16.9 }  }$} 45.8 {\tiny$\xrightarrow{  \mathbf{ +\ \ 5.3 }  }$} 51.1 & 57.7 {\tiny$\xrightarrow{  \mathbf{ +19.0 }  }$} 76.7 {\tiny$\xrightarrow{  \mathbf{ +\ \ 4.4 }  }$} 81.1 & 38.7 {\tiny$\xrightarrow{  \mathbf{ +17.8 }  }$} 56.5 {\tiny$\xrightarrow{  \mathbf{ +\ \ 5.0 }  }$} 61.5\\
 & Soft (min.) & 39.4 {\tiny$\xrightarrow{  \mathbf{ +11.6 }  }$} 51.0 {\tiny$\xrightarrow{  \mathbf{ +\ \ 0.6 }  }$} \textbf{51.6} & 67.4 {\tiny$\xrightarrow{  \mathbf{ +14.0 }  }$} 81.4 {\tiny$\xrightarrow{  \mathbf{ +\ \ 0.5 }  }$} \textbf{81.9} & 49.1 {\tiny$\xrightarrow{  \mathbf{ +12.5 }  }$} 61.6 {\tiny$\xrightarrow{  \mathbf{ +\ \ 0.5 }  }$} \textbf{62.1}\\
 & Soft (par.) & 39.4 {\tiny$\xrightarrow{  \mathbf{ +\ \ 9.2 }  }$} 48.6 {\tiny$\xrightarrow{  \mathbf{ +\ \ 2.5 }  }$} 51.1 & 67.4 {\tiny$\xrightarrow{  \mathbf{ +12.6 }  }$} 80.0 {\tiny$\xrightarrow{  \mathbf{ +\ \ 1.7 }  }$} 81.7 & 49.1 {\tiny$\xrightarrow{  \mathbf{ +10.5 }  }$} 59.6 {\tiny$\xrightarrow{  \mathbf{ +\ \ 2.1 }  }$} 61.7\\
 & Soft (ran.) & \ \ 2.3 {\tiny$\xrightarrow{  \mathbf{ +47.1 }  }$} 49.4 {\tiny$\xrightarrow{  \mathbf{ +\ \ 1.9 }  }$} 51.3 & \ \ 8.0 {\tiny$\xrightarrow{  \mathbf{ +73.0 }  }$} 81.0 {\tiny$\xrightarrow{  \mathbf{ +\ \ 0.7 }  }$} 81.7 & \ \ 4.5 {\tiny$\xrightarrow{  \mathbf{ +55.9 }  }$} 60.4 {\tiny$\xrightarrow{  \mathbf{ +\ \ 1.5 }  }$} 61.9\\
 \hline
\multirow{3}{*}{Rob} & \textsc{lpaqa} & \ \ 1.2\reimp & \ \ 9.1\reimp & \ \ 4.2\reimp\\
 & AutoPrompt & 40.0 & 68.3 & 49.9\\
 & Soft (min.) & \ \ 1.2 {\tiny$\xrightarrow{  \mathbf{ +39.4 }  }$} \textbf{40.6} {\tiny$\xrightarrow{  \mathbf{ -\ \ 7.3 }  }$} 33.2 & \ \ 9.1 {\tiny$\xrightarrow{  \mathbf{ +66.3 }  }$} \textbf{75.4} {\tiny$\xrightarrow{  \mathbf{ -22.3 }  }$} 53.0 & \ \ 4.2 {\tiny$\xrightarrow{  \mathbf{ +48.8 }  }$} \textbf{53.0} {\tiny$\xrightarrow{  \mathbf{ -12.1 }  }$} 40.8\\
 \hline
\multirow{2}{*}{BAb} & \textsc{lpaqa} & \ \ 0.8\reimp &\ \ 5.7\reimp & \ \ 2.9\reimp\\
 & Soft (min.) & \ \ 0.8 {\tiny$\xrightarrow{  \mathbf{ +39.1 }  }$} \textbf{39.9} &\ \ 5.7 {\tiny$\xrightarrow{  \mathbf{ +69.7 }  }$} \textbf{75.4} & \ \ 2.9 {\tiny$\xrightarrow{  \mathbf{ +49.2 }  }$} \textbf{52.1}\\
 \hline
\multirow{2}{*}{BAl} & \textsc{lpaqa} & \ \ 3.5\reimp &\ \ 5.6\reimp & \ \ 4.8\reimp\\
 & Soft (min.) & \ \ 3.5 {\tiny$\xrightarrow{  \mathbf{ +22.3 }  }$} \textbf{25.8} & \ \ 5.6 {\tiny$\xrightarrow{  \mathbf{ +62.4 }  }$} \textbf{68.0} & \ \ 4.8 {\tiny$\xrightarrow{  \mathbf{ +36.2 }  }$} \textbf{41.0} \\
\specialrule{.1em}{.05em}{.05em}
\end{tabular}
\caption{
\label{tab:trex-ori-full}
Experimental results on T-REx original datasets. In the LM column, BEb is BERT-base-cased, BEl is BERT-large-cased, BAb is BART-base-cased, BAl is BART-large-cased, Rob is RoBERTa-base, and Rol is RoBERTa-large.
In the results block, ``init'' uses the initial untuned prompts; ``soft'' starts at ``init'' and tunes the prompts (layer 0) and mixture weights; and ``deep'' starts at ``init'' and tunes all the layers.
Numbers above the arrows are the relative change in the performance.
Within each block, we boldface the best system and all those that are not
significantly worse (paired permutation test, $p < 0.02$).
We also boldface the relative changes that are significantly different from 0.
Other symbols are as in \cref{tab:trex}.
}
\end{table*}

\begin{table*}[t]
\centering
\begin{tabular}{@{\hspace{0.2\tabcolsep}}c@{\hspace{0.2\tabcolsep}}|@{\hspace{0.2\tabcolsep}}c@{\hspace{0.2\tabcolsep}}@{\hspace{0.2\tabcolsep}}|@{\hspace{0.2\tabcolsep}}l@{\hspace{0.2\tabcolsep}}|@{\hspace{0.2\tabcolsep}}l@{\hspace{0.2\tabcolsep}}|@{\hspace{0.2\tabcolsep}}l}
\specialrule{.1em}{.05em}{.05em}
\multirow{2}{*}{LM} & \multirow{2}{*}{Method} & \multicolumn{1}{c}{Precision@1} & \multicolumn{1}{c}{Precision@10} & \multicolumn{1}{c}{MRR}\\
\cline{3-5}
 &  & \ init\ \ \ \,$\rightarrow$\ \ \ \,soft\ \ \ \,$\rightarrow$\ \ \ \,deep & \ init\ \ \ \,$\rightarrow$\ \ \ \,soft\ \ \ \,$\rightarrow$\ \ \ \,deep & \ init\ \ \ \,$\rightarrow$\ \ \ \,soft\ \ \ \,$\rightarrow$\ \ \ \,deep\\
 \hline
\multirow{6}{*}{BEb} & \textsc{lama} & 26.4 & 54.3 & 35.8\\
 & \textsc{lpaqa} & 31.2 & 57.3 & 39.9\\
 & Soft (sin.) & 26.4 {\tiny$\xrightarrow{  \mathbf{ +22.2^? }  }$} 48.6 {\tiny$\xrightarrow{  \mathbf{ +\ \ 1.0 }  }$} 49.6 & 54.3 {\tiny$\xrightarrow{  \mathbf{ +23.3^? }  }$} 77.6 {\tiny$\xrightarrow{  \mathbf{ +\ \ 0.3 }  }$} 77.9 & 35.8 {\tiny$\xrightarrow{  \mathbf{ +22.9^? }  }$} 58.7 {\tiny$\xrightarrow{  \mathbf{ +\ \ 0.6 }  }$} 59.3\\
 & Soft (min.) & 31.2 {\tiny$\xrightarrow{  \mathbf{ +19.0^? }  }$} 50.2 {\tiny$\xrightarrow{  \mathbf{ +\ \ 0.3 }  }$} \textbf{50.5}\notcomparable & 57.3 {\tiny$\xrightarrow{  \mathbf{ +21.9^? }  }$} 79.2 {\tiny$\xrightarrow{  \mathbf{ +\ \ 0.5 }  }$} \textbf{79.7}\notcomparable & 39.9 {\tiny$\xrightarrow{  \mathbf{ +20.2^? }  }$} 60.1 {\tiny$\xrightarrow{  \mathbf{ +\ \ 0.4 }  }$} \textbf{60.5}\notcomparable\\
 & Soft (par.) & 31.2 {\tiny$\xrightarrow{  \mathbf{ +18.5^? }  }$} 49.7 {\tiny$\xrightarrow{  +\ \ 0.0 }$} 49.7 & 57.3 {\tiny$\xrightarrow{  \mathbf{ +21.3^? }  }$} 78.6 {\tiny$\xrightarrow{  \mathbf{ +\ \ 0.6 }  }$} 79.2 & 39.9 {\tiny$\xrightarrow{  \mathbf{ +19.6^? }  }$} 59.5 {\tiny$\xrightarrow{  \mathbf{ +\ \ 0.3 }  }$} 59.8\\
 & Soft (ran.) & \ \ 0.8 {\tiny$\xrightarrow{  \mathbf{ +46.3\ \ \ }  }$} 47.1 {\tiny$\xrightarrow{  \mathbf{ +\ \ 3.5 }  }$} 50.6 & \ \ 4.0 {\tiny$\xrightarrow{  \mathbf{ +70.4\ \ \ }  }$} 74.4 {\tiny$\xrightarrow{  \mathbf{ +\ \ 4.9 }  }$} 79.3 & \ \ 2.2 {\tiny$\xrightarrow{  \mathbf{ +54.3 \ \ \ }  }$} 56.5 {\tiny$\xrightarrow{  \mathbf{ +\ \ 3.9 }  }$} 60.4\\
 \hline
\multirow{6}{*}{BEl} & \textsc{lama} & 24.0\reimp & 53.7\reimp & 34.1\reimp\\
 & \textsc{lpaqa} & 37.8\reimp & 64.4\reimp & 44.0\reimp\\
 & Soft (sin.) & 24.0 {\tiny$\xrightarrow{  \mathbf{ +26.2 }  }$} 50.2 {\tiny$\xrightarrow{  \mathbf{ +\ \ 1.2 }  }$} 51.4 & 53.7 {\tiny$\xrightarrow{  \mathbf{ +24.9 }  }$} 78.6 {\tiny$\xrightarrow{  \mathbf{ +\ \ 0.9 }  }$} 79.5 & 34.1 {\tiny$\xrightarrow{  \mathbf{ +25.9 }  }$} 60.0 {\tiny$\xrightarrow{  \mathbf{ +\ \ 1.2 }  }$} 61.2\\
 & Soft (min.) & 37.8 {\tiny$\xrightarrow{  \mathbf{ +13.4 }  }$} 51.2 {\tiny$\xrightarrow{  \mathbf{ +\ \ 1.3 }  }$} \textbf{52.5} & 64.4 {\tiny$\xrightarrow{  \mathbf{ +15.1 }  }$} 79.5 {\tiny$\xrightarrow{  \mathbf{ +\ \ 1.6 }  }$} \textbf{81.1} & 44.0 {\tiny$\xrightarrow{  \mathbf{ +17.0 }  }$} 61.0 {\tiny$\xrightarrow{  \mathbf{ +\ \ 1.4 }  }$} \textbf{62.4}\\
 & Soft (par.) & 37.8 {\tiny$\xrightarrow{  \mathbf{ +12.5 }  }$} 50.3 {\tiny$\xrightarrow{  \mathbf{ +\ \ 1.4 }  }$} 51.7 & 64.4 {\tiny$\xrightarrow{  \mathbf{ +14.3 }  }$} 78.7 {\tiny$\xrightarrow{  \mathbf{ +\ \ 2.1 }  }$} 80.8 & 44.0 {\tiny$\xrightarrow{  \mathbf{ +16.1 }  }$} 60.1 {\tiny$\xrightarrow{  \mathbf{ +\ \ 1.6 }  }$} 61.7\\
 & Soft (ran.) & \ \ 1.4 {\tiny$\xrightarrow{  \mathbf{ +46.1 }  }$} 47.5 {\tiny$\xrightarrow{  \mathbf{ +\ \ 4.4 }  }$} 51.9 & \ \ 5.4 {\tiny$\xrightarrow{  \mathbf{ +68.9 }  }$} 74.3 {\tiny$\xrightarrow{  \mathbf{ +\ \ 6.3 }  }$} 80.6 & \ \ 5.7 {\tiny$\xrightarrow{  \mathbf{ +51.2 }  }$} 56.9 {\tiny$\xrightarrow{  \mathbf{ +\ \ 5.0 }  }$} 61.9\\
\specialrule{.1em}{.05em}{.05em}
\end{tabular}
\caption{
\label{tab:trex-ext-full}
Experiment results on T-REx extended datasets.
}
\end{table*}

\begin{table*}[hbt!]
\centering
\begin{tabular}{c|cccccccccccc}
\specialrule{.1em}{.05em}{.05em}
prompts & T-REx-min. & T-REx-par. & Goog-sin. & Goog-min. & Goog-par. & ConceptNet\\
\hline
\#relations   & 41 & 41 & 3 & 3 & 3 & 16  \\
avg. prompts  & 28.4 & 26.2 & 1 & 32.7 & 28.0 & 9.3 \\
min \#prompts & 6 & 13 & 1 & 29 &24 & 1 \\
max \#prompts & 29 & 30 & 1 & 40 & 30 & 38 \\
avg. \#tokens & 5.1 & 4.5 & 4.7 & 5.3 & 4.2 & 7.1 \\
\specialrule{.1em}{.05em}{.05em}
\end{tabular}
\caption{\label{tab:prompt_stats}
Statistics of prompts.
The ``Goog'' stands for ``Google-RE.''
We do not list the statistics of randomized prompts, as
they should match the statistics of the mined prompts (``min.'') from
which they are derived.
}
\end{table*}

\begin{table*}[h]
\centering
\begin{tabular}{c|cccccccccccc}
\specialrule{.1em}{.05em}{.05em}
database & T-REx original & T-REx extended & Google-RE & ConceptNet\\
\hline
\#relations   & 41 & 41 & 3 & 16   \\
avg. \#unique $\px$  & 1580 & 834 & 1837 & 511 \\
avg. \#unique $\py$ & 217 & 151 & 372 & 507 \\
min \#$(\px, \py)$ & 544 & 310 & 766 & 510 \\
max \#$(\px, \py)$ & 1982 & 1000 & 2937 & 4000\\
mean \#$(\px, \py)$ & 1715 & 885 & 1843 & 1861  \\
\specialrule{.1em}{.05em}{.05em}
\end{tabular}
\caption{\label{tab:rel_stats}
Statistics of the relational databases.}
\end{table*}

\,\\

\begin{table}[t]
\centering
\begin{tabular}{@{\hspace{0.0\tabcolsep}}c|l@{\hspace{0.5\tabcolsep}}l@{\hspace{0.5\tabcolsep}}l@{\hspace{0.0\tabcolsep}}}
\specialrule{.1em}{.05em}{.05em}
Model & \multicolumn{1}{c}{P@1} & \multicolumn{1}{c}{P@10} & \multicolumn{1}{c}{MRR}\\
\specialrule{.1em}{.05em}{.05em}
\textsc{lpaqa} (BEb) & 18.9 & 40.4 & 26.6 \\
Soft (BEb) & \textbf{23.0} {\small (+\textbf{4.1})} & \textbf{45.2} {\small (+\textbf{4.8})} & \textbf{30.5} {\small (+\textbf{3.9})} \\
\hline
\textsc{lpaqa} (BEl) & 23.8 & 47.7 & 32.2 \\
Soft (BEl) & \textbf{27.0} {\small (+\textbf{3.2})} & \textbf{51.7} {\small (+\textbf{4.0})} & \textbf{35.4} {\small (+\textbf{3.2})} \\
\specialrule{.1em}{.05em}{.05em}
\end{tabular}
\caption{
\label{tab:distinct}
Results with distinct $y$'s.
We use the BERT-base-cased and BERT-large-cased LMs and the LPAQA mining based prompts as initial prompts.
The experiments are conducted on the T-REx original dataset.
}%
\end{table}

\begin{table}[t]
    \centering
    \begin{tabular}{c|cccccc}
\specialrule{.1em}{.05em}{.05em}
        statistic & mean & std &min & max  \\
        \hline
        T-REx original + min. & 12.5 & 4.0 & 4.6 & 21.0 \\
        T-REx extended + min. & 12.5 & 4.0& 4.6 & 20.3  \\
        T-REx original + par. & 5.4 & 4.0 & 1.1 & 17.1  \\
        T-REx extended + par. & 5.4 & 3.9 & 1.2 & 18.4 \\
\specialrule{.1em}{.05em}{.05em}
    \end{tabular}
    \caption{Statistics of effective number of prompts.}
    \label{tab:eff_prompts}
\end{table}

\end{document}